\documentclass[preprint,3p,11pt]{elsarticle}

\usepackage{amsmath}
\usepackage{amssymb}
\usepackage{hyperref}
\usepackage{natbib}
\journal{Elsevier}

\begin{document}

\begin{frontmatter}



\title{Surrogate modeling for stochastic crack growth processes in structural health monitoring applications}


\author[NTUA]{Nicholas E. Silionis}
\author[NTUA]{Konstantinos N. Anyfantis\texorpdfstring{\corref{cor}}{}}

\affiliation[NTUA]{organization={Ship-Hull Structural Health Monitoring (S-H SHM) Group, School of Naval Architecture and Marine Engineering, National Technical University of Athens},%Department and Organization
            addressline={9 Heroon Polytechniou Av.}, 
            city={Athens},
            postcode={15780 Zografos}, 
            country={Greece}}

\ead{kanyf@naval.ntua.gr}
\cortext[cor]{Corresponding author. Tel.: +30 210 772 1325}

\begin{abstract}
Fatigue crack growth is one of the most common types of deterioration in metal structures with significant implications on their reliability. Recent advances in Structural Health Monitoring (SHM) have motivated the use of structural response data to predict future crack growth under uncertainty, in order to enable a transition towards predictive maintenance. Accurately representing different sources of uncertainty in stochastic crack growth (SCG) processes is a non-trivial task. The present work builds on previous research on physics-based SCG modeling under both material and load-related uncertainty. The aim here is to construct computationally efficient, probabilistic surrogate models for SCG processes that successfully encode these different sources of uncertainty. An approach inspired by latent variable modeling is employed that utilizes Gaussian Process (GP) regression models to enable the surrogates to be used to generate prior distributions for different Bayesian SHM tasks as the application of interest. Implementation is carried out in a numerical setting and model performance is assessed for two fundamental crack SHM problems; namely crack length monitoring (damage quantification) and crack growth monitoring (damage prognosis).
\end{abstract}

%% \begin{graphicalabstract}
%% \end{graphicalabstract}

%% \begin{highlights}
%% \item Research highlight 1
%% \item Research highlight 2
%% \end{highlights}


\end{frontmatter}


\section{Introduction} \label{Seq1}

A significant number of structures across such diverse fields as marine and offshore, aerospace, civil infrastructure etc., are currently operating at or close to the limit of their design life. As they are subject to different deterioration phenomena, it is inevitable that some operate at decreased levels of performance and consequently increased levels of risk with regard to failure. One of the most common types of deterioration, for metal structures in particular, are cracks that propagate under the influence of dynamic loading. In certain domains, such as ship structures which are the authors' main focus, it is expected that fatigue cracks exist in structures currently in operation, especially aging ones. Current monitoring practices typically specify a series of on-site inspection events, based on either visual assessment or more specialized techniques like liquid penetrant tests \citep{iacsrec84}. The emergence of the field of Structural Health Monitoring (SHM) has shifted interest towards predictive or condition-based maintenance (CBM), where the goal is to use data from in-situ sensors to continuously monitor deteriorating structures and enable safe lifetime extension as well as flexible maintenance planning through prognostic models \citep{Jardine2006, Thelen_v1_2022, Thelen_v2_2022, Silionis_v1_2023}.

On account of their prevalence and importance to structural safety, fatigue cracks have been a focal point of SHM research \citep{Yao2014} with works dealing with all different levels of the SHM hierarchy proposed by Rytter \citep{rytter1993}; namely, damage detection \citep{Carden2004, Alvandi2006, Cholevas2023}, localization \citep{Shokrani2018, Tatsis2020, Argyris2018}, quantification \citep{Sbarufatti2013, Tatsis2022, Oboe2023} and prognosis \citep{Corbetta2018, Cristiani2021, Kamariotis2023}. Naturally, the prognostic aspect of SHM is of particular interest when it comes to fatigue crack growth \citep{Leser2020,Galanopoulos2023}, as its focus is to obtain probabilistic predictions of the evolution of structural deterioration. These can be in turn updated in the presence of data, and therefore offer more robust predictions on quantities of interest such as the remaining useful life (RUL) or the probability of failure. Using data to update prior knowledge on particular aspects of crack growth processes offers a natural means to quantify the uncertainty associated with them, which has long been recognized by the research community \citep{Ray1996,CROSS2007} and is primarily associated with material-related properties and fatigue loading \citep{Kamariotis2023, Straub2009, Giannella2021, Giannella2022, DiFrancesco2022, Makris2023}.

The present work is concerned with the effect of uncertainty on crack growth processes that are described using crack growth models based on fracture mechanics (e.g., Paris-Erdogan \citep{Paris1963} or NASGRO \citep{NASGRO}). Typical approaches for uncertainty quantification (UQ) in crack growth models are based on sampling techniques belonging to the broader family of Monte Carlo (MC) methods \citep{Giannella2021,Giannella2022}. These rely on first assuming a specific probabilistic model over the different parameters that are considered as random, i.e., material, load-related or both, and subsequently obtaining realizations of the crack growth process by propagating samples drawn from the probability distributions of the parameters to the crack growth model. Since such models operate under the assumption of constant fatigue loading during crack propagation, these methods have to be modified in case the structure of interest is subject to operational conditions characterized by varying levels of stress amplitude and frequency. Such conditions are however commonly encountered by numerous structural systems today, including ocean going vessels and wind turbines among others \citep{Li2013,Qi2018,Mylonas2021}.

\begin{figure}[t!]
	\centering
	\includegraphics[scale=0.9]{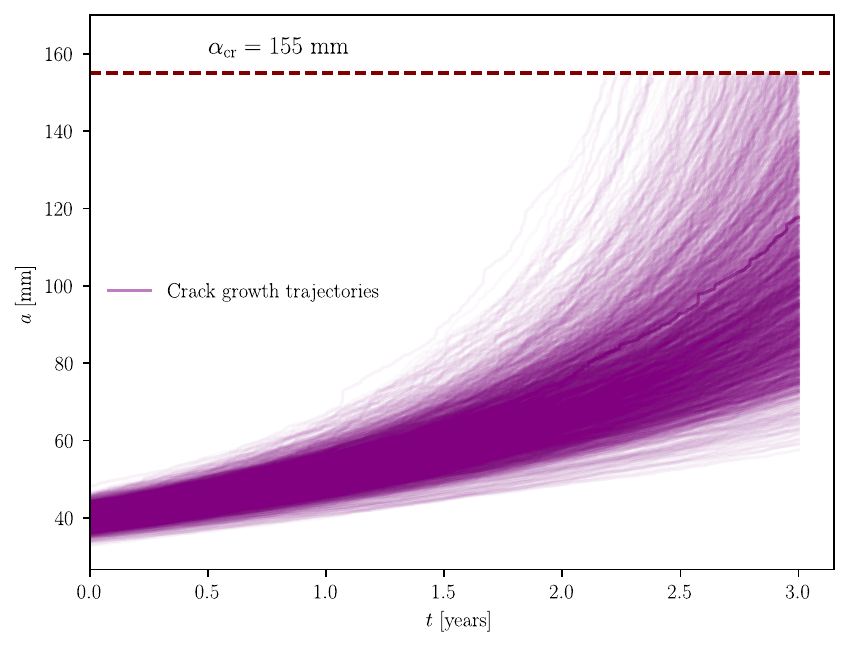}
	\caption{Realizations of a stochastic crack growth process according to Makris et al. \citep{Makris2023}}
	\label{Fig 1}
	
\end{figure}

In a previous work focusing on ship structures \citep{Makris2023}, the authors developed a numerical scheme that uses MC simulation to propagate uncertainty over the crack growth model parameters and simultaneously employs a time-discretization algorithm to account for load variability in each crack growth realization. Although the original goal was to generate RUL distributions and time-based reliability estimates, the outcomes of the numerical scheme effectively represent realizations of a stochastic crack growth process, as shown in Figure \ref{Fig 1}. These constitute a valuable resource for both diagnostic and prognostic fatigue crack monitoring systems that follow the model-based, and more specifically Bayesian paradigm, as they can be employed to construct physically consistent prior distributions.

The general goal of Bayesian methods for SHM is to use structural response measurements to obtain up-to-date estimates on the probability distribution over a set of structural parameters that are of monitoring interest, e.g., crack length at some point in time or the time-invariant parameters of a crack growth model. This is achieved by first assuming a prior distribution over those parameters and then using Bayes' rule to obtain a data-informed posterior distribution (see \citep{Simoen2013} for a more detailed description). Selecting an appropriate prior distribution can lead to more rapid convergence to the posterior distribution, and in some cases even to an analytical derivation \citep{Gelman2013}; more pertinently, if chosen judiciously it can ensure that updated quantities remain consistent with underlying problem physics. The crack growth trajectories demonstrated in Figure \ref{Fig 1} correspond to realizations from a physically consistent crack growth prior process; they are not however in their initial form suitable to construct Bayesian priors as they require a computationally costly, physics-based simulation to be obtained.

The objective of the present work is to propose an efficient, unified framework that utilizes Gaussian Process (GP) regression models to learn a parsimonious surrogate model of the prior process. Moreover, by exploiting principles of latent variable modeling, the proposed methodology is capable of constructing priors suited to different SHM tasks, such as crack length monitoring (quantification) and crack growth monitoring (prognosis). By virtue of the employed model, these priors are able to effectively encode uncertainty related both to time-invariant (material-related), and time-variant (load-related) parameters. To the authors' knowledge, this is the first time that such a unified, physically consistent approach to Bayesian prior construction for crack SHM is presented in the literature. Furthermore, through its use of probabilistic Machine Learning (ML) tools, the proposed surrogate demonstrates a practical and computationally efficient framework that can be applied within the general sphere of fatigue crack growth problems to account for the stochasticity that is inherent to them. 

A brief, yet self-contained, description of the process employed to generate stochastic crack growth trajectories following the work of Makris et al. \citep{Makris2023} is presented in Section \ref{Seq2}. Section \ref{Seq3} introduces Gaussian Process regression, with a focus on sparse methods that can deal with large quantities of data. Priors constructed using the proposed model for different SHM tasks are presented in Section \ref{Seq4}, while Section \ref{Seq 5} offers concluding remarks.

\section{Modeling stochastic crack growth trajectories} \label{Seq2}

The dataset employed throughout this work was obtained using a methodology proposed by the authors \citep{Makris2023}, where a spectral method was combined with a stochastic crack growth model to generate SCG process realizations, to be referred heretofore as trajectories, that are representative of a typical marine structural element under realistic operational conditions. In this work, only a brief outline of this process will be provided and the primary focus will be placed on the characteristics of the trajectories themselves and how these informed the choice of the proposed surrogate modeling framework. For more detailed information the interested reader is referred to the original work \citep{Makris2023}.

To generate the crack growth trajectories shown in Figure \ref{Fig 1}, it was initially assumed that a crack already exists on a particular structural component (double-bottom girder) of a containership, whose length at the time of detection was considered as a Gaussian random variable, i.e., $\alpha_0 \sim \mathcal{N}(\mu_{\alpha_0}, \sigma^2_{\alpha_0})$. This is consistent with commonly accepted industry opinion that fatigue cracks exist on operating ships and that when detected, some uncertainty is expected in the initial measurement due to equipment-related noise which is commonly modeled using a Gaussian distribution \citep{BS2015}. To model the crack growth process, the Paris-Erdogan law \citep{Paris1963} was employed, which in the time domain has the following general form:

\begin{equation} \label{eq2_1}
	\frac{d \alpha}{dt} = N_{\text{avg}} C \left[ \Delta K (\Delta S) \right]^{m}
\end{equation}

which holds when the fatigue loading pair $\{ \Delta S, N_{\text{avg}} \}$ is time-invariant. However, fatigue loading of ship hull structural components is primarily caused by ocean waves, which are stationary only over specific periods of time, known as sea states \citep{Massel1996}. In order to use the Paris-Erdogan law, a discretization of the operational lifetime of the vessel was employed to obtain intervals that correspond to individual sea states. For each of these, a set of parameters describing its intensity, i.e., significant wave height $H_{\text{S}}$, and duration, i.e., zero up-crossing period $T_{\text{Z}}$, were sampled from a joint probability model based on oceanographic observations \citep{ABS2022}. These were then transformed under a linear model for ship hydrodynamics to the corresponding fatigue loading pairs $\{ \Delta S, N_{\text{avg}} \}$, thus allowing the model of Eq. (\ref{eq2_1}) to be used within each temporal interval.

To account for material-related uncertainty, the time-invariant model parameters $C,m$ were assumed to be random variables. According to commonly accepted practice in the literature (e.g., \citep{Kamariotis2023}), they were considered to follow the lognormal and normal distributions respectively; namely $C \sim \ln \mathcal{N}(\mu_{C}, \sigma^2_{C})$ and $m \sim \mathcal{N}(\mu_{m}, \sigma^2_{m})$. A closed form equation for the stress intensity factor $\Delta K$ was employed as the structural component in question can be considered to operate, without loss of generality, as a plate. Details on the numerical values assigned to these quantities are provided in the original work \citep{Makris2023} and are not included here in the interest of brevity.

To generate the trajectories depicted in Figure \ref{Fig 1} a MC simulation was first applied over the model-related parameters and the initial crack length to generate $N=10^4$ tuples, namely $\left\lbrace C^{(i)},m^{(i)},\alpha_0^{(i)} \right\rbrace_{i=1}^{N}$, each of which corresponds to a specific trajectory. The time-discretization scheme was then employed for each one by assuming a uniformly distributed random sea-state duration, ranging from 5-7 hours. The discontinuity caused by this discretization is evident in the trajectories, which generally exhibit a non-smooth behavior. The duration of the crack propagation was restricted to 3 years after the initial crack was detected, in a choice motivated by the 5-year fixed inspection schedule followed by ships \citep{IACS2021}. Furthermore, an upper limit to the crack length was set at $\alpha_{\text{cr}} = 155 \ \text{mm}$, according to a failure criterion established based on a structural reliability analysis (see Makris et al. \citep{Makris2023}). As a result of the constraints imposed on the numerical scheme, individual trajectories ultimately do not have the same length, i.e., number of points. This is an interesting characteristic of the data, that prohibits the use of traditional batch regression techniques that are based on learning one-to-one mappings \citep{Rasmussen2005}; it can also be interpreted as a reflection of the underlying dynamics of the crack growth process.

The dataset obtained from the initial physics-based simulation has a prohibitively large size, due to the sea state duration, which is itself dictated by physical considerations. Such a fine discretization offers no added value in terms of adequately capturing the underlying probabilistic structure of the crack growth process. Therefore, a subsampling scheme was implemented to reduce the dimensionality of the dataset, while retaining the probabilistic structure of the data. According to this, it was considered that for each trajectory, the crack length is available at bimonthly intervals. This allows for the essential characteristic of different trajectory length to be retained; it also ensures that the resultant crack length values correspond to the same points in time. The latter transforms the data to a format consistent with the theoretical principles underpinning stochastic processes, which allows estimating mean and covariance functions and broadens the scope of tools available for analysis.

\begin{figure}[b!]
	\centering
	\includegraphics[scale=0.85]{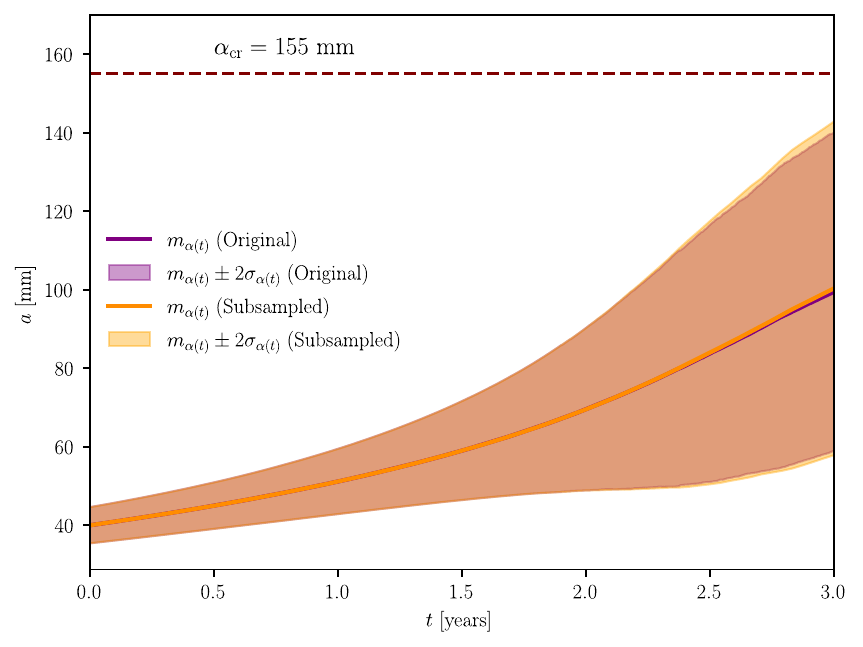}
	\caption{Comparison between initial and reduced dataset after subsampling. The darker orange color of the 95 \% C.I. is a result of the overlap between the orange and purple colors corresponding to each dataset.}
	\label{Fig 2}
	
\end{figure}

In Figure \ref{Fig 2}, a comparative depiction of the mean functions and $95 \%$ credible intervals (C.I.) for the original and subsampled data is provided; evidently, the reduced dataset sufficiently captures the probabilistic characteristics of the crack growth process. Not only is the mean trend captured but so is the heteroscedastic, i.e., input-dependent, variance structure; the latter is another key characteristic of the data, and also poses a potential challenge in choosing the surrogate model form. Based on the preceding analysis and stated goal of this work, this form has to be such that it is unaffected by the crack growth trajectory length and can also model heterescodastic behavior. In order to be used effectively to construct priors for different SHM tasks, it must also be able to generate conditional distributions over the crack length at different levels of conditioning variables; namely time instances, crack growth model parameters and the initial crack length. Finally, it has to be functional for large numbers of data, as even after subsampling the dataset employed in this work has considerable size.

Gaussian Process regression models \citep{Rasmussen2005} have been chosen as the surrogate here, as they satisfy all these requirements under certain conditions. They are by definition able to provide probabilistic predictions over outputs, i.e., crack lengths, and are able to accommodate potentially multi-dimensional inputs; thus they are able to provide the flexibility required for conditioning on different variables. Being framed on the assumption that all outputs belong to an underlying Gaussian Process, they operate in a point-wise fashion and thus are unaffected by different trajectory lengths. In their sparse formulation they can accommodate large datasets and also model heteroscedastic behavior \citep{Snelson2005}. They can also be trained effectively with the help of variational learning \citep{Titsias2009,Hensman2013,Hensman2015}, while offering an inherent guarantee against overfitting \citep{Rasmussen2001}. What follows is a description of the basic theoretical underpinnings of GP regression, with a focus on sparse GPs and variational learning.
 
\section{Gaussian Process regression} \label{Seq3}
\subsection{Fundamental principles} \label{SubSeq3.1}

We begin with the typical problem setting for supervised learning; namely let us consider that there exist $N$ available training observations, arranged in a training set $\mathcal{D} = \{ \left( \mathbf{x}_i, y_i \right) \}_{i=1}^{N}$, where $\mathbf{x}_i  \in \mathbb{R}^d$ are $d$-dimensional inputs, while $y_i \in \mathbb{R}$ are noisy observations of an unobserved or latent function $f(\mathbf{x}_i)$. If we consider the noise to be independent Gaussian with variance $\sigma^2$, then for every noisy observation in the dataset:

\begin{equation} \label{eq1}
	y_i = f(\mathbf{x}_i) + \epsilon_i, \qquad \epsilon_i \sim \mathcal{N}(0, \sigma^2)
\end{equation}

The goal of a GP is to essentially set a prior over functions $f$ and then use Bayesian inference to obtain the posterior distribution in light of the observed data \cite{Rasmussen2005}. This prior can be expressed as:

\begin{equation} \label{eq2}
	f(\mathbf{x}) \sim \mathcal{GP}(m_f(\mathbf{x}; \theta), k_f(\mathbf{x}, \mathbf{x'}; \theta	))
\end{equation}

and is fully defined by its mean and covariance functions, $m_f(\cdot)$ and $k_f(\cdot, \cdot)$ respectively, which in turn are controlled by a vector of hyperparameters $\theta$, the dependency on which will be heretofore implied and removed from the notation. To obtain the posterior over $f$, Bayes' rule is applied:

\begin{equation} \label{eq3}
p(\mathbf{f}|\mathbf{y}) = \frac{p(\mathbf{y}| \mathbf{f})p(\mathbf{f})}{\int p(\mathbf{y}| \mathbf{f})p(\mathbf{f}) \,d\mathbf{f}}
\end{equation}

where $p(\mathbf{f})$ is the GP prior and $p(\mathbf{y} | \mathbf{f})$ is the likelihood function, which for the observation model we have considered is also Gaussian. Furthermore, taking into account the stochastic independence between observations, we can write:

\begin{equation}\label{eq4}
p(\mathbf{y} | \mathbf{f}) = \prod_{i=1}^{N} \mathcal{N}(\mathbf{m}_f, \mathbf{K}_f + \sigma^2\mathbb{I}_{N})
\end{equation}

where $\mathbf{m}_f$ and $\mathbf{K}_f$ refer to entries to the mean vector and covariance matrix, defined as:

\begin{equation}\label{eq5}
\mathbf{m}_f[i] \triangleq m_f(\mathbf{x}_i)
\end{equation}

\begin{equation}\label{eq6}
\mathbf{K}_f[i,j] \triangleq k_f(\mathbf{x}_i, \mathbf{x}_j)
\end{equation}

By virtue of of Eq. (\ref{eq2}) \& (\ref{eq4}), the posterior distribution of Eq. (\ref{eq3}) is also Gaussian with tractable mean and covariance functions:

\begin{equation}\label{eq7}
m_{\mathbf{y}}(\mathbf{x}) = K_{\mathbf{x}n}(\sigma^2\mathbb{I}_{N} + K_{nn})^{-1}\mathbf{y}
\end{equation}

\begin{equation}\label{eq8}
k_{\mathbf{y}}(\mathbf{x}, \mathbf{x'}) = k_f(\mathbf{x}, \mathbf{x'}) - K_{\mathbf{x}n}(\sigma^2\mathbb{I}_{N} + K_{nn})^{-1}K_{n\mathbf{x'}}
\end{equation}

Where $K_{nn}$ is the $N \times N$ covariance matrix on the training inputs, $K_{\mathbf{x}n}$ is an $N$-dimensional row vector of covariance function values between $\mathbf{x}$ and the training inputs,  $K_{n\mathbf{x'}} = K^{\top}_{\mathbf{x}n}$ and $\mathbb{I}_{N}$ is the $N$-dimensional identity matrix. This form of tractable posterior leads also to a tractable posterior predictive distribution, which describes the probability of obtaining a prediction $y_{*}$ at some unseen input $\mathbf{x}_{*}$, and is given by:

\begin{equation}\label{eq9}
p(y_{*}|\mathbf{y}) = \int p(\mathbf{y_{*}}|\mathbf{f})p(\mathbf{f}|\mathbf{y}) \,d\mathbf{f} = \mathcal{N}(y_{*}|m_{\mathbf{y}}(\mathbf{x_{*}}), k_f(\mathbf{x_{*}}, \mathbf{x_{*}}) + \sigma^2)
\end{equation}

The posterior GP, and by extension the posterior predictive distribution, depend on the values of the hyperparameters of the mean and covariance functions and the observation noise standard deviation, namely $\{ \theta, \sigma^2 \}$. Training a GP consists of estimating them so as to maximize the logarithm of the marginal likelihood, which in this case is also tractable and given by:

\begin{equation}\label{eq10}
\log p(\mathbf{y}) = \log \left[ \mathcal{N}(\mathbf{y}|\mathbf{0}, \sigma^2\mathbb{I}_{N} + K_{nn}) \right]
\end{equation}

Clearly, the expressions contained in Eq. (\ref{eq10}) are unsuitable for large datasets, since they include the inversion of an $N \times N$ matrix which scales as $\mathcal{O}(N^3)$. By definition, they are also incapable of modeling a heteroscedastic, i.e., input-dependent, variance structure due to the independent Gaussian assumption for the observation noise. The latter is desirable for the problem at hand, given the nature of the crack growth data.

\subsection{Variational learning for sparse Gaussian Process regression} \label{SubSeq3.2}

In this work, inducing point methods will be employed which are used to construct Sparse Gaussian Processes (SGPs) and are suitable for treating large datasets as well as modeling heteroscedastic variance, albeit to a certain extent \citep{Snelson2005}. Inducing point methods rely on introducing a set of inducing variables $\mathbf{u}=\{f(\mathbf{z}_i)\}_{i=1}^{M}, M \ll N$, that are points of the function calculated at the inducing points $Z = \{\mathbf{z}_i)\}_{i=1}^{M}$ and therefore belong to the same space as $\mathbf{x}$. Since the inducing variables also belong to the same space as $\mathbf{f}$, then $p(\mathbf{f}, \mathbf{u})$ is jointly Gaussian. Ultimately, the goal is to use the inducing variables to directly approximate the posterior GP mean and covariance functions of Eq. (\ref{eq7}) \& (\ref{eq8}) at significantly lower computational cost. Constructing this approximation consists of selecting both the inducing points as well as the model hyperparameters so as to maximize the log-marginal likelihood $p(\mathbf{y})$, at a reduced computational cost and while retaining the Gaussianity of the posterior \citep{Titsias2009}.

Variational Inference (VI) will employed for this task following the example of Hensman et al. \citep{Hensman2013,Hensman2015}; the goal of VI is to approximate the true posterior using another distribution, known as the variational distribution, by framing the problem as one of optimization \citep{Blei2017}. In this context, the parameters of the variational distribution act as design variables while the objective function is a measure of distance between the two distributions. The set of parameters that minimizes this distance yields a distribution that is very similar to the true posterior. Here, we shall denote the variational distribution as $q(\mathbf{f})$; the variational objective is to minimize the Kullback-Leibler (KL) divergence $\mathrm{KL}\left( q(\mathbf{f})||p(\mathbf{f}|\mathbf{y}) \right)$ between the two distributions, which is a well-known metric of similarity between probability density functions \citep{Kullback1951}. By expanding the KL divergence (see Blei et al. \citep{Blei2017} for a more detailed proof), the following expression can be obtained:

\begin{equation}\label{eq11}
\mathrm{KL}\left( q(\mathbf{f})||p(\mathbf{f}|\mathbf{y}) \right) = \mathrm{KL}\left(q(\mathbf{f})||p(\mathbf{f})\right) + \log p(\mathbf{y}) - \mathbb{E}_{q(\mathbf{f})}\left[ \log p(\mathbf{y}|\mathbf{f}) \right]
\end{equation}

where it is observed that the variational objective is connected to the evidence term that we seek to, but cannot effectively, maximize. However, noting that the KL divergence is by definition non-negative we obtain:

\begin{equation}\label{eq12}
\log p(\mathbf{y}) \geq \mathbb{E}_{q(\mathbf{f})}\left[ \log p(\mathbf{y}|\mathbf{f}) \right] -\mathrm{KL}\left(q(\mathbf{f})||p(\mathbf{f})\right)   
\end{equation}

The term on the right-hand side provides a lower bound on the log-marginal likelihood and is known as the Evidence Lower Bound (ELBO). Contrary to the evidence term itself, the ELBO is a more convenient objective function as it is more amenable to gradient-based optimization. Initially, the inducing variables do not appear in the ELBO. However, recalling that they belong to the same space as function outputs, then $q(\mathbf{f})$ can be written as $q(\mathbf{f}) = \int p(\mathbf{f}|\mathbf{u})q(\mathbf{u}) \, d\mathbf{u}$. The typical assumption of a Gaussian variational distribution over the inducing variables  is then made  \citep{Hensman2015}, i.e., $q(\mathbf{u}) = \mathcal{N}(\mathbf{m}, \mathbf{S})$, where $\mathbf{m}$ and $\mathbf{S}$ denote the mean vector and covariance matrix respectively. It follows that the approximate posterior is available in functional form:

\begin{equation}\label{eq13}
q(\mathbf{f}) = \mathcal{N}\left( \mathbf{f} | \mathbf{A}\mathbf{m}, \mathbf{K}_{nn} + \mathbf{A} \left( \mathbf{S} - \mathbf{K}_{mm} \right)\mathbf{A}^{\top} \right)
\end{equation}

where $\mathbf{A} = \mathbf{K}_{nn}\mathbf{K}_{mm}^{-1}$ and $\mathbf{K}_{mm}$ is the $M \times M$ covariance matrix on the inducing points, which is now the only matrix subject to inversion. The last step in order to complete deriving the ELBO, which can then be maximized with respect to $\{ \mathbf{m}, \mathbf{S} \}$ is to use the stochastic independence of the observations to factorize the likelihood function term, which yields:

\begin{equation}\label{eq14}
\mathcal{L}_{ELBO} = \sum_{n=1}^{N}\mathbb{E}_{q(f_n)}\left[ \log p(y_n \mid f_n) \right] - \mathrm{KL}\left[ q(\mathbf{u}) \vert \vert p(\mathbf{u})) \right]
\end{equation}

In this form, we can now use gradient based optimization to find the parameters of $q(\mathbf{u})$ that maximize this bound on the log-marginal likelihood. Using the optimal parameters, predictions at some unseen test input $\mathbf{x}_{*}$ and corresponding latent function value $f_{*}$ can be made using the following predictive equation:

\begin{equation}\label{eq15}
p(f_{*}|\mathbf{y}) = \int p(f_{*}\vert\mathbf{u}) q(\mathbf{u}) \,d\mathbf{u}
\end{equation}

This integral is also tractable and the mean and variance of $f_{*}$ can be calculated, from which follows the calculation of the posterior predictive distribution $p(y_{*} \vert \mathbf{y})$ as in Eq. (\ref{eq9}).

\section{Surrogate modeling for stochastic crack growth} \label{Seq4}

The proposed surrogate model will be implemented and showcased for prior distribution construction on two fundamental crack SHM tasks, i.e., crack length and crack growth monitoring. Training the model and assessing its performance will be discussed and results will be presented for the two tasks, along with an investigation on the effect of adding varying levels of knowledge to the model through different conditioning variables.

\subsection{Building the GP regression model} \label{SubSeq4.1}

Building a GP regression model starts from specifying the characteristics of the GP prior (see Eq. (\ref{eq2})) in the form of its mean and covariance functions. In principle, the mean function can be any function of the inputs, but is typically set to zero because of the relative flexibility of GPs to model complex relationships \citep{Rasmussen2005}, especially for data which have been transformed to a zero-mean space. Accordingly, a zero mean function has been selected here as well. The role of the covariance (or kernel) function is to control the level of similarity between pairs of input points, which is ultimately reflected in the covariance matrix, whose entries are defined as in Eq. (\ref{eq6}). Several popular kernel choices exist for the covariance function; in this work we have employed a Mat\'ern 3/2 kernel, which is defined as follows:

\begin{equation} \label{eq17}
k_{f} \left( \mathbf{x}_{i}, \mathbf{x}_{j} \right) = \alpha^2 \left( 1 + \frac{\sqrt{3} \lvert \mathbf{x}_{i} - \mathbf{x}_{j} \rvert}{l} \right) \exp \left( - \frac{\sqrt{3} \lvert \mathbf{x}_{i} - \mathbf{x}_{j} \rvert}{l} \right)
\end{equation}

where $\theta = \lbrace \alpha, l \rbrace$ are the kernel function hyperparameters; the process variance $\alpha$ controls variations around the mean while the length scale $l$ represents the smoothness of the function \citep{Bull2023}. In the conventional setting, training GPs with a zero mean function consists of finding the set of hyperparameters $\theta$ that maximize the log-marginal likelihood. For sparse GPs, training includes simultaneously learning the inducing point locations as well. The ELBO, as described in the previous sections, offers a convenient objective function formulation that accommodates both sets of parameters, along with those related to the variational distribution. Moreover, VI allows for state-of-the-art stochastic gradient descent algorithms to be used as the engine of the training process.

For all applications this work is concerned with, a mean-field Gaussian approximation was employed as the variational distribution \citep{Blei2017}, while the Adam optimizer \citep{Kingma2014} was used to train the model. The number of iterations for training as well as the learning rate were adjusted for each model to ensure optimal performance. A hold-out scheme was employed to construct the training set, as well as the test set used to assess model performance. According to this, an equal portion of the initial 10$^4$ crack growth trajectories was assigned to each set, which due to the unequal number of points in each curve led to slightly unbalanced training and test sets. Trajectories were sampled randomly so as to ensure that no bias occurs during selection.

The ELBO defined in Eq. (\ref{eq14}) was used to monitor model performance over the training set. Subsequently, the trained model  fit was assessed using two different metrics. The first was the normalized mean square error (NMSE), defined here as in Rogers et al. \citep{Rogers2020}:

\begin{equation}\label{eq18}
NMSE = \frac{100}{N_{\text{test}} \sigma^2_y} \sqrt{\left( \mathbf{y} - \mathbf{\hat{y}} \right)^{\top}\left( \mathbf{y} - \mathbf{\hat{y}} \right)}
\end{equation}

where $N_{\text{test}}$ refers to the number of samples in the test set, which are collected in the vector $\mathbf{y}$ and have variance $\sigma^2_{y}$; finally, the model predictions over the test set are denoted as $\mathbf{\hat{y}}$. The NMSE captures the predictive capacity of the model in terms of point predictions, therefore for probabilistic models, such as the one employed herein, it can be used to assess their accuracy in terms of the mean prediction. In the way it is defined here, a perfect fit returns a score of zero while a score of 100 refers to a model with the same predictive capacity as simply taking the mean over the data. Since uncertainty quantification is both a strength as well as one of the main motivators behind model selection in this work, it is desirable to employ a performance metric that is capable of providing a probabilistic outlook on the goodness-of-fit.

Such a metric can be found in the GP formulation itself, more specifically in the form of the joint likelihood function described in Eq. (\ref{eq4}). This provides a direct means to assign a likelihood to each model prediction based on the commonly employed zero-mean Gaussian prediction error formulation \citep{Kennedy2001}. For strictly computational purposes, this work employs the logarithm of the likelihood as the metric, as it conveniently transforms the product in the joint likelihood to a sum. Crucially, it must be noted that to ensure physical consistency is maintained when assessing the model, both metrics are evaluated on a trajectory-to-trajectory basis. 

\subsection{Constructing priors for crack length monitoring} \label{SubSeq4.2}

The surrogate model is first implemented for the task of constructing prior distributions for the problem of crack length monitoring. This corresponds to a typical damage quantification setting, which in the Bayesian context requires choosing a prior over the crack length at a given moment in time, i.e., $p \left( \alpha (t) \right)$. This can be obtained as the output of the surrogate model by considering a training set $\mathcal{D}_{\text{I}} = \lbrace \left( \mathbf{x}_i, y_i \right) \rbrace_{i=1}^{N} = \left \lbrace \left( t^{(i)}, a^{(i)} \right) \right \rbrace_{i=1}^{N}$, where $t_i, \alpha_i \in \mathbb{R}$. In this setting, all sources of uncertainty affecting the crack length, namely the initial crack length as well as material- and load-related variability, are treated as latent variables of the surrogate model. Training took place over 1000 iterations requiring approximately 7 minutes in a workstation equipped with an NVIDIA\textsuperscript{\textregistered} RTX A4000 GPU. Implementation took place using the Pyro probabilistic programming language written in Python \citep{Pyro}.

\begin{figure}[htp!]
	\centering
	\includegraphics[scale=0.85]{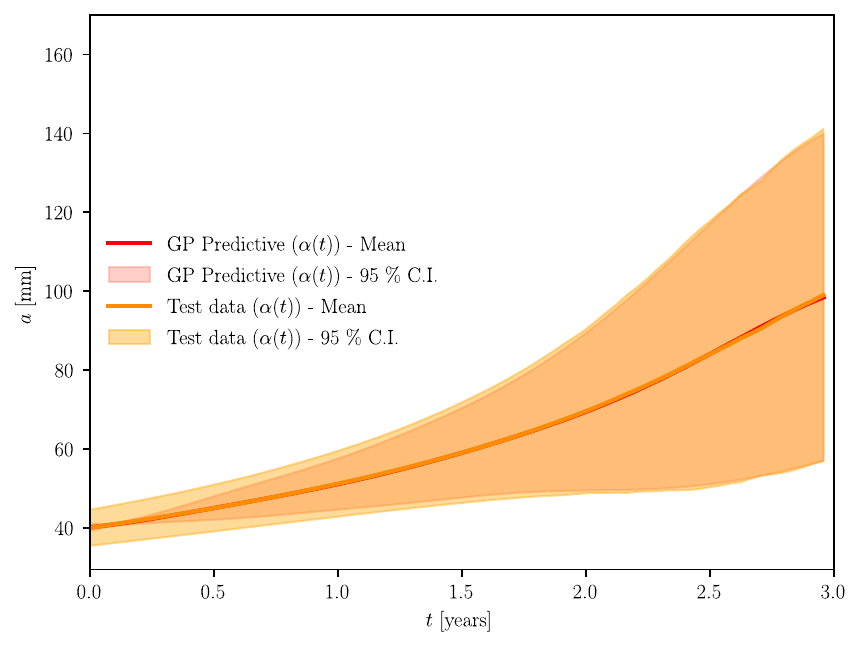}
	\caption{Surrogate model predictive performance for crack length prior construction}
	\label{Fig 3}
	
\end{figure}

The predictive performance of the trained model is showcased in Figure \ref{Fig 3}, where the mean function is plotted over the 3 year crack growth period alongside the 95 \% C.I. obtained using the covariance function. This is compared with the mean function estimated over the test data and the corresponding credible interval. The model is capable of precisely capturing the mean trend in the data, while also being largely successful in representing the heteroscedasic variance structure. However, it is noteworthy that the predicted variance underestimates the actual variance of the test data early in the crack propagation process. This indicates an inability of the model to successfully account for the uncertainty in the initial condition, i.e., the initial crack length $\alpha_0$. 

The observed variance depletion could pose a problem in a Bayesian setting, as insufficient variance in the prior could lead to poor convergence to the posterior. However, as the crack growth rate increases this phenomenon subsides, which indicates that the generated priors will be sufficient as deterioration becomes more extensive, and thus structural reliability is decreased. It should be noted that the NMSE and log-likelihood for this model will be reported in the next section in comparison to the model trained for crack growth monitoring, as it was felt that this was more contextually appropriate.

\cleardoublepage

\subsection{Constructing priors for crack growth monitoring} \label{SubSeq4.3}

Crack growth monitoring using Bayesian inference consists of using available measurements, either of the structural response or of the crack length itself, to obtain estimates of the posterior distribution over the crack growth model parameters, namely $C,m$ in the Paris-Erdogan law (see Eq. (\ref{eq2_1})). The probabilistic nature of these parameters is well-attested in the literature \citep{Kamariotis2023,Giannella2022} and suggested priors were used in the authors' previous work \citep{Makris2023} to generate the SCG trajectories shown in Figure \ref{Fig 1}. To apply Bayesian inference, a likelihood function is required which is capable of performing a transformation from the prior (parameter) space to the observable (measurement) space. Regardless of the measured quantity, a probabilistic model is required which returns a distribution over the crack length at some moment in time, conditioned the parameters $C,m$, i.e., $p \left( \alpha(t) \vert C,m \right)$.

Albeit providing realizations from this underlying distribution, the employed numerical scheme used to generate crack growth trajectories does not provide a convenient model for prior construction. The proposed surrogate can achieve precisely that by being trained over a training set $\mathcal{D}_{\text{II}} = \lbrace\left( \mathbf{x}_i, y_i \right) \rbrace_{i=1}^{N} =  \left \lbrace \left( \left[ t^{(i)} \ C^{(i)} \ m^{(i)} \right]^{\top}, a_i \right) \right\rbrace_{i=1}^{N}$, where now $\mathbf{x}_i \in \mathbb{R}^3$ and $\alpha_i \in \mathbb{R}$. Under this parametrization, model parameters are included in the set of observable variables while load-related ones and the initial crack length are passed onto the latent space.  Again, training took place  for 1000 iterations on the same workstation requiring a similar amount of time.

\begin{figure}[t!]
	\centering
	\includegraphics[scale=0.85]{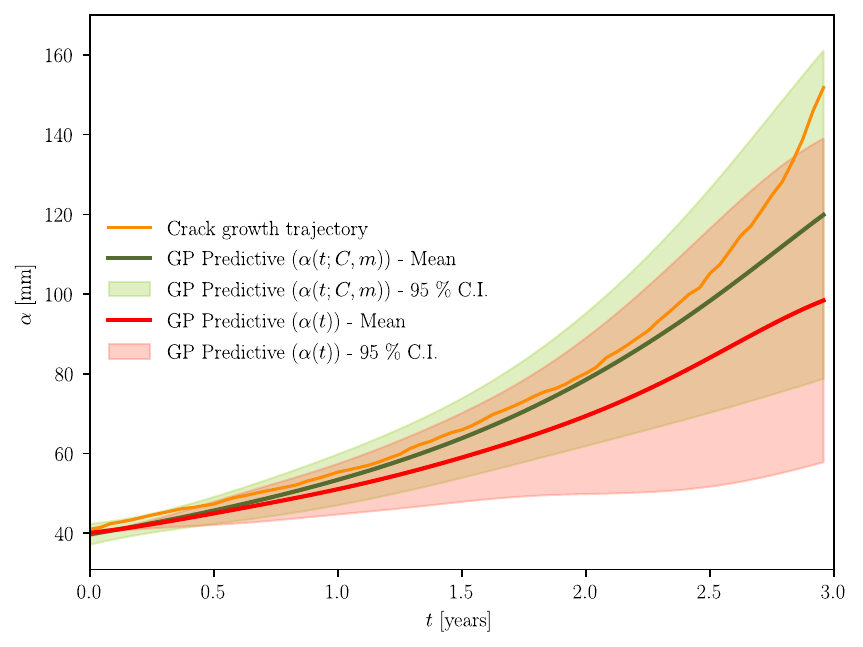}
	\caption{Comparison between prior models for crack length and crack growth monitoring with uncertain initial crack length}
	\label{Fig 4}
	
\end{figure}

The effect of conditioning on the crack growth model parameters is evident in Figure \ref{Fig 4}. There, the mean function and 95 \% C.I. of the predictive crack growth process $\alpha\left(t; C,m \right)$ are plotted for a specific $C,m$ realization and compared against the actual crack growth trajectory as well as the mean and 95 \% C.I. of $\alpha(t)$ from Section \ref{SubSeq4.2}. This particular trajectory is representative of the effect of the stochastic sequence of fatigue loads, as it exhibits non-smooth behavior, as well as that of outlying parameter realizations that lead to a rapidly increasing crack growth rate. As a result, the model that treats all sources of uncertainty as latent variables is singificantly outperformed by that conditioned on $C, m$. This is further reinforced by the fact that the log-likelihood of the actual trajectory with respect to the model of the $\alpha\left(t; C,m \right)$ process is seven times greater than the one with respect to $\alpha(t)$.

However, in absolute terms that likelihood is still negative, which is reflective of the fact that for significant portions of the crack growth the actual trajectory falls outside of the high probability density region predicted by the model. Nevertheless, it should be noted that the additional parametrization leads to marked uncertainty reduction, along with a more robust description of the heteroscedastic tendency of the data. This is especially evident from the fact that the 95 \% C.I. of the GP predictive process for $\alpha\left(t; C,m \right)$ extends to crack lengths at the limit of the critical threshold ($\alpha_{\text{cr}} = 155 \ \text{mm}$), which is not the case for $\alpha(t)$.

\begin{figure}[t!]
	\centering
	\includegraphics[scale=0.62]{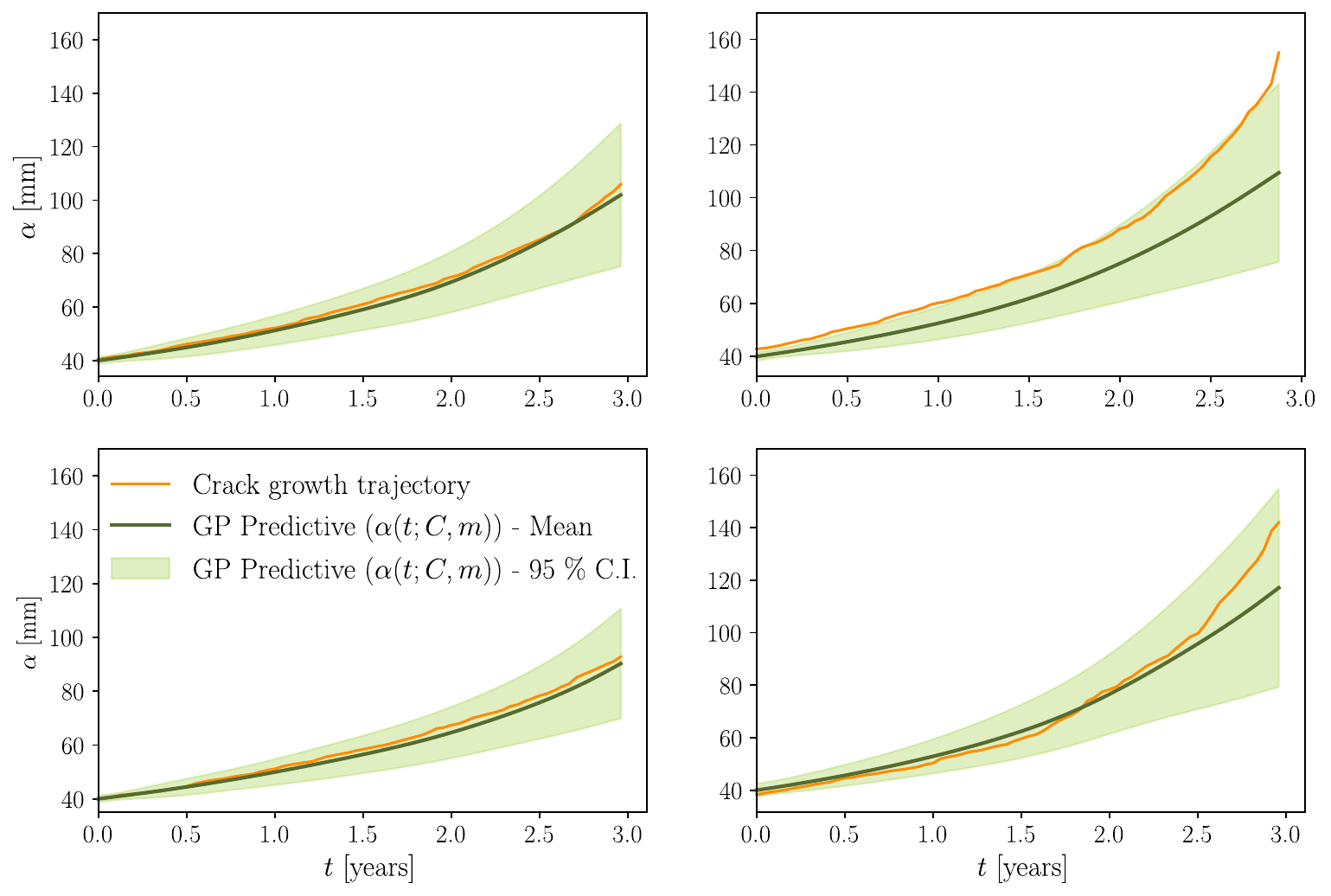}
	\caption{Predictive performance of prior model for crack growth monitoring for different $C,m$ realizations}
	\label{Fig 5}
	
\end{figure}

The predictive capacity of the trained GP for $\alpha\left(t; C,m \right)$ is presented more extensively in Figure \ref{Fig 5}. To produce it, four previously unseen crack growth trajectories were selected randomly from the test set. The predictive mean and 95 \% C.I. were obtained from the trained surrogate for the corresponding $C,m$ realizations and the results are plotted comparatively. The trajectories depicted on the left panels are typical of the mean behavior of the SCG process, as they exhibit relatively moderate crack growth rates and somewhat smoother shapes. The model in these cases is highly effective in capturing the mean trend while also providing a narrow high probability density region.

In the top-right panel, the actual trajectory begins from an outlying initial crack length and then exhibits a rapid increase in crack growth rate beyond approximately the 1.5 year mark. This is indicative of the effect of the stochastic time-variant loading, as more severe sea states and therefore $\lbrace \Delta S, N_{\text{avg}} \rbrace$ pairs, lead to acceleration of crack growth. As a result, the model largely fails to provide a sufficient prediction for prior construction. Interestingly, while the case in the bottom-right panel also exhibits the mark of significant uncertainty in the loading sequence, the predictive performance is visibly improved. This is a consequence of the fact that the initial crack length for the actual trajectory is not an outlier, thus highlighting the effect of uncertainty in the initial crack length.

Motivated by this, we decided to investigate the predictive performance of the surrogate when the initial crack length is included in the conditioning variables. This decision also has a practical dimension, in that when monitoring crack growth an initial crack is expected to have been detected and its length measured, up to some confidence level. The training set now becomes $\mathcal{D}_{\text{III}} = \lbrace\left( \mathbf{x}_i, y_i \right) \rbrace_{i=1}^{N} =  \left \lbrace \left( \left[ t^{(i)} \ C^{(i)} \ m^{(i)} \ \alpha^{(i)}_0 \right]^{\top}, a_i \right) \right\rbrace_{i=1}^{N}$, with $\mathbf{x}_i \in \mathbb{R}^4$ and $\alpha_i \in \mathbb{R}$. This time, training was implemented over 2000 iterations which required approximately 14 minutes using the same machine, as well as hold-out scheme for training and test set construction. The resultant GP predictive process for $\alpha\left(t; C,m,\alpha_0 \right)$ features only fatigue loading-related quantities as its latent variables.

\begin{figure}[htp!]
	\centering
	\includegraphics[scale=0.85]{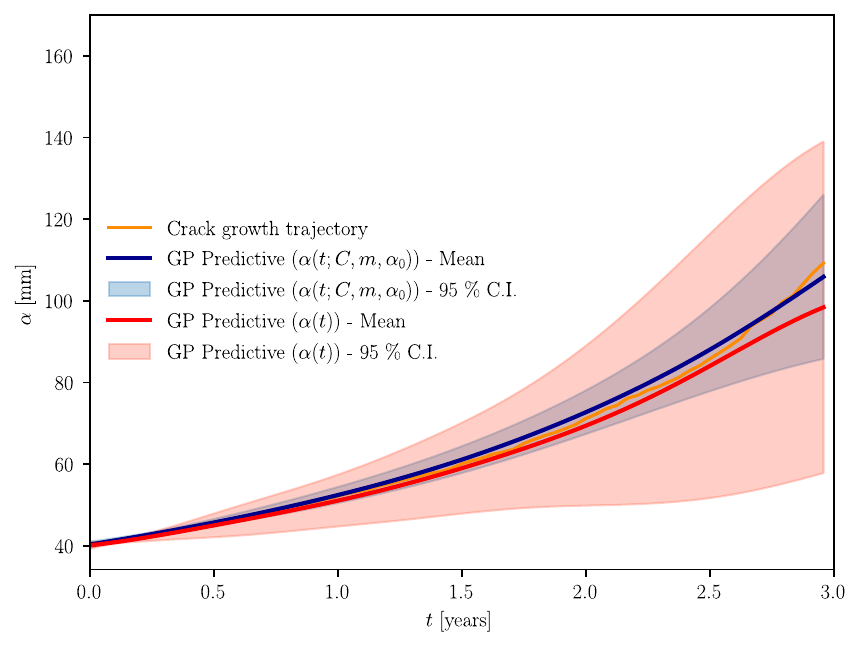}
	\caption{Comparison between prior models for crack length and crack growth monitoring including conditioning on the initial crack length $\alpha_0$}
	\label{Fig 6}
\end{figure}

\begin{figure}[t!]
	\centering
	\includegraphics[scale=0.85]{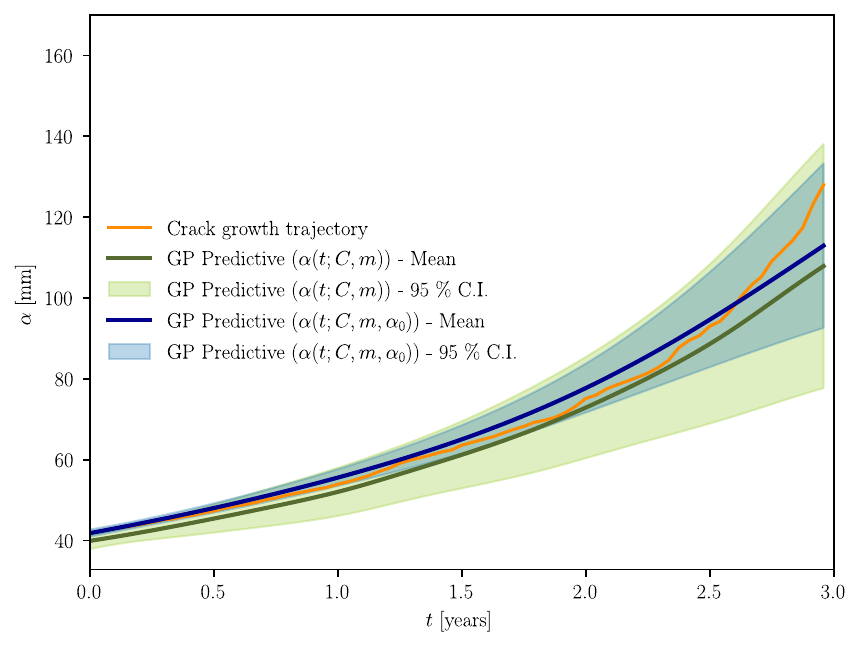}
	\caption{Comparison between prior models for crack growth monitoring with and without conditioning on initial crack length $\alpha_0$ }
	\label{Fig 7}
	
\end{figure}

The uncertainty reduction enabled by the inclusion of the initial crack length to the model inputs is clearly demonstrated in Figure \ref{Fig 6}, where the predicted 95 \% C.I. is drastically narrower. A fourfold increase is observed in the log-likelihood of the trajectory, along with an NMSE decrease from 0.2 to 0.1; the very low values signifying the overall success of the model in terms of capturing the mean trend. Furthermore, its capacity to capture heteroscedastic behavior has clearly been improved; this is further reinforced by the results presented in Figure \ref{Fig 7}.

There, despite the fact that the initial crack length for the actual trajectory is an outlier, the model is capable of capturing the crack growth very effectively, providing an almost point-wise accurate prediction over the first year. After that, a relative lag is observed alongside an increasingly less smooth behavior which indicates that less severe loading has caused a deceleration of the crack growth process. This is followed by a rapid increase in the crack growth rate indicating a reversal in the fatigue loading pattern. The trained model is proven capable of following this stochastic behavior by producing a highly heteroscedastic predictive process which, compared to the predictive process for $\alpha \left( t; C,m \right)$, exhibits lower variance throughout.

\begin{figure}[t!]
	\centering
	\includegraphics[scale=0.60]{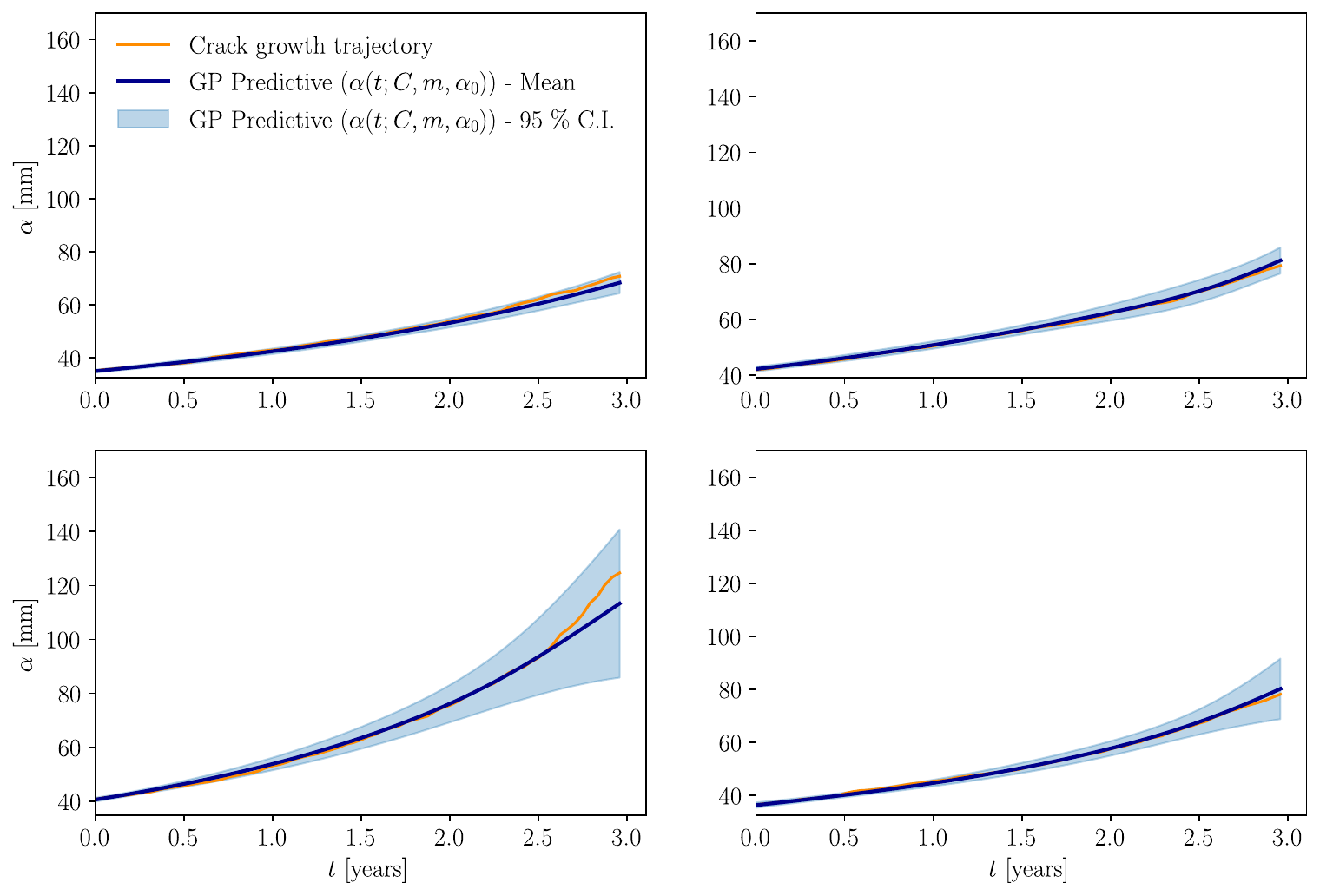}
	\caption{Predictive performance of prior model for crack growth monitoring for different $C,m,\alpha_0$ realizations}
	\label{Fig 8}
	
\end{figure}

Quantitatively, the log-likelihood of the crack growth trajectory is three times greater for the GP predictive process $\alpha \left( t; C,m,\alpha_0 \right)$, compared to $\alpha \left( t; C,m \right)$, with respective NMSE values of 0.08 and 0.11. While relatively equivalent in terms of mean predictions, adding knowledge to the model about the initial conditions has a decidedly positive effect on reducing uncertainty, as is also made clear from Figure \ref{Fig 8}. The crack growth trajectories and corresponding model predictions shown therein were obtained for test set realizations of both the Paris-Erdogan parameters $C,m$ and the initial crack length $\alpha_0$.

The GP predictive processes for the top panels, which contain smooth trajectories largely unaffected by loading variability, exhibit almost point-wise consistency in the mean alongside very narrow credible intervals. For the bottom panels where loading variability is more pronounced, especially on the left-hand side, the model still produces heteroscedastic processes that accurately capture the trends in the actual trajectories. As expected, the performance of the model predictive processes is unaffected by outlying initial crack lengths, such as the one on the top-left panel.

\begin{figure}[htp!]
	\centering
	\includegraphics[scale=0.78]{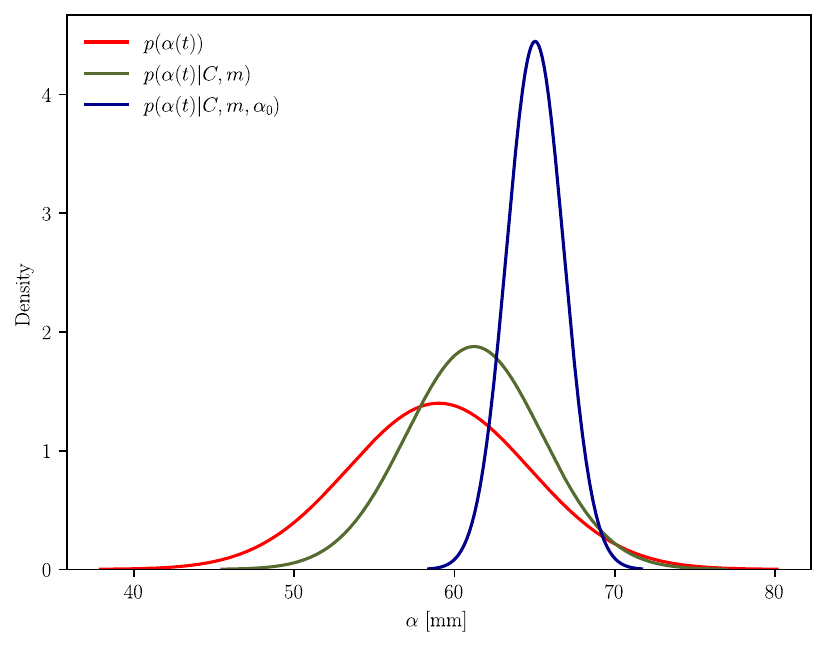}
	\caption{Crack length priors under different levels of knowledge about the crack growth process at $t=1.5$ years}
	\label{Fig 9}
	
\end{figure}

Figure \ref{Fig 9} further showcases the capability of the model to generate prior distributions over the crack length at different moments in time and under different levels of conditioning variables. For $t=1.5$ years, the trained predictive GP models are used to generate (Gaussian) prior distributions. For the models including conditioning variables, i.e., $\alpha \left( t; C,m \right)$ and $\alpha \left( t; C,m,\alpha_0 \right)$, the same realizations of $C,m,\alpha_0$ as in Figures \ref{Fig 6} \& \ref{Fig 7} were employed. Here as well, the uncertainty reduction is evident, while the variance of the resulting distributions is seen to increase under lower levels of information, i.e., when more variables are treated as latent. This leads to wider priors which are less prone, in the Bayesian setting, to introduce bias during the inference process. Therefore, in addition to its ability to accurately represent problem physics, as demonstrated previously, the model is shown to exhibit  statistical properties which prove it is suitable for the task of constructing physically consistent Bayesian priors.

\cleardoublepage

\section{Concluding Remarks} \label{Seq 5}

This work demonstrated the potential of using GP regression models as surrogates for stochastic crack growth processes in order to construct physically consistent prior distributions in Bayesian SHM problems. The proposed model is capable of accounting for different sources of uncertainty, either directly as input parameters or indirectly as latent variables, thus allowing it to be used under different levels of physical knowledge as well as for a hierarchy of different tasks. Implementation took place using an existing dataset of crack growth realizations for a typical ship structural component, which were obtained using the Paris-Erdogan law and taking into account both material and load-related uncertainty.

When used to construct prior distributions for crack length monitoring, where all uncertain quantities are modeled as latent variables, the surrogate was proven up to the task for most of the crack growth duration. However, in the initial phase of crack growth where lower growth rates are observed the model was found to underestimate process variance, which could prove problematic in the Bayesian setting. Although generally capable of modeling heteroscedastic behavior due to the use of inducing point methods, the employed model is not strictly built to model heteroscedastic variance. This could provide an avenue for future research in view of improving model predictive performance for the crack length monitoring task.

The introduction of more knowledge on the physical parameters of the crack growth process, in the form of the initial crack length and/or the Paris-Erdogan law parameters, led to a reduction in the predictive uncertainty of the model, as well an improvement of its capacity to model heteroscedastic variance. This observation provides incentive for further research into the impact of introducing physical knowledge into the model, this time not through the training data but through its structure. Such a grey-box modeling approach, where problem physics can be introduced either through the mean function shape or via constraints on the covariance function, offers the possibility for a model that can generalize well using smaller amounts of potentially lower quality, and thus cheaper to obtain, data.

Finally, it is important to state that this work constitutes the first step in a broader research effort that the authors are currently undertaking. The developed models are meant to act as parts of a hierarchical methodology that aims to tackle both tasks the models were demonstrated on, i.e., crack length and crack growth monitoring, in a simultaneous and interchangeable manner.

\section*{Acknowledgements}

The authors would like to gratefully acknowledge the contribution of Pavlos Makris, who was instrumental in producing the stochastic crack growth dataset employed throughout this work.




% \bibliographystyle{elsarticle-num}

\bibliography{Int_J_Fatigue_Ref}


%% \begin{thebibliography}{00}


%% \bibitem{}

%% \end{thebibliography}
\end{document}